# Using an Improved Output Feedback MPC Approach for Developing a Haptic Virtual Training System

**Soroush Sadeghnejad*[1], Farshad Khadivar[2], Mojtaba Esfandiari[2], Golchehr Amirkhani[2], Hamed Moradi[2], Farzam Farahmand[2] and Gholamreza Vossoughi[2]**

[1]Biomedical Engineering Department, Amirkabir University of Technology (Tehran Polytechnic). No. 350, Hafez Ave, Valiasr Square, Tehran 1591634311, I.R. IRAN.

[2]School of Mechanical Engineering, Sharif University of Technology. Azadi Ave., Tehran 1458889694, I.R. IRAN.

*Corresponding Author
Assistant Professor, Biomedical Engineering Department, Amirkabir University of Technology (Tehran Polytechnic)
Email: s.sadeghnejad@aut.ac.ir
Tel: +98 (21) 64545952; Fax: +98 (21) 66468186

**ABSTRACT** Haptic training simulators generally consist of three major components, namely a human operator, a haptic interface, and a virtual environment. Appropriate dynamic modeling of each of these components can have far-reaching implications for the whole system's performance improvement in terms of transparency, the analogy to the real environment, and stability. In this paper, we developed a virtual-based haptic training simulator for Endoscopic Sinus Surgery (ESS) by doing a dynamic characterization of the phenomenological sinus tissue fracture in the virtual environment, using an input-constrained linear parametric variable model. A parallel robot manipulator equipped with a calibrated force sensor is employed as a haptic interface. A lumped five-parameter single-degree-of-freedom mass-stiffness-damping impedance model is assigned to the operator's arm dynamic. A robust online output feedback quasi-min-max model predictive control (MPC) framework is proposed to stabilize the system during the switching between the piecewise linear dynamics of the virtual environment. The simulations and the experimental results demonstrate the effectiveness of the proposed control algorithm in terms of robustness and convergence to the desired impedance quantities.

**Keywords:** Virtual reality-based haptic system, Endoscopic Sinus Surgery (ESS) training simulator, Linear parametric variable, Model predictive control (MPC), Robust stability, Quasi-min–max algorithm.

1. Introduction

Nowadays, endoscopic surgeries have grown in popularity among surgeons for the treatment of low-grade and high-grade tumors located in the nasal cavities, the paranasal sinus, and the anterior skull base [40, 13]. To become a dexterous surgeon in endoscopic device manipulation and hand-eye coordination, one needs to undergo special training to be able to safely perform an efficacious endoscopic sinus and skull base surgery (ESSS) [41]. The virtual reality-based haptic training systems development provides medical students and residents with a practically effective training framework, on which they can practice and learn various surgical operations in virtual environments simulating real operations to them. As a result, surgeons can be adept in several surgical techniques in advance of performing real endoscopic interventions simply by practicing on these surgical simulator systems [34, 3].

Each virtual reality (VR) based haptic simulator system involves three main parts: the human operator, the haptic interface, and the virtual environment. An ideal VR-based haptic simulator provides the operator with the capability to interact with the simulated virtual environment in a stable and transparent fashion without a noticeable time delay [15, 16]. The development of VR-based haptic simulator systems seems attractive from a practical point of view since it can address challenging surgical tasks, such as dynamic interactions between the surgical endoscopic device and the virtual environment. These challenges emanate from the transition between the diverse virtual environments with different impedances such as soft and hard tissues or sudden tissue fracture [1, 22, 24], the nonlinearity of the dynamics of the (parallel) haptic interface, and the human operator's arm dynamics [20]. The presence of such nonlinear dynamics and the resultant uncertainties demand proper dynamic modeling as well as an efficient control scheme design [21].

A haptic interface is an essential component of the VR simulators of endoscopic surgeries [29]. Therefore, the control framework of the haptic interface, for example, unilateral, bilateral, or multilateral control schemes, should be designed such that it enables stable interaction with either virtual environments in simulators or remote environments in teleoperated systems. Closed-loop system stability is studied by considering the unmodeled dynamic terms and the uncertainties in the models of the human operator and the virtual environment, using the linear time-invariant (LTI) differential models [12] and the passive models [10, 11]. Quite a few control strategies have been utilized for the VR haptic simulator and teleoperation systems so far [12, 38]. Passivity-based algorithms [36], wave theories [39], robust control approaches [7, 9, 19], adaptive methods [17-19], impedance control [14], force reflecting control [28], and optimal control were introduced in recent years.

Considering the previous control frameworks, an alternative proposed control scheme for tackling the input, state, and output constraints as well as compensating for the negative effects of communication channel time delay is model predictive control approaches [35, 38]. The use of Model Predictive Control (MPC) algorithms has grown recently due to their capability to deal with the linear and nonlinear multi-input multi-output (MIMO) dynamic systems with uncertain time-delay, switching phenomenon, and constraints on the inputs, outputs, and states [8, 18, 20], especially in the presence of uncertain parameters as well as external disturbances [43]. Uddin et. al. surveyed a systematic report of the up-to-date model predictive control algorithms and their applications in bilateral teleoperation systems [42].

Sirouspour and Shahdi represented the mathematical formulas for a time-delay teleoperation system as a multimodal continuous-time linear–quadratic–Gaussian (LQG) synthesis model [37]. To enhance the adaptability and performance of a bilateral teleoperation framework, De Rossi et al. [4] employed the Smith predictor with an adaptive buffering scheme to lower the system's sensitivity to the time-varying delays, probable packet losses, and noise and disturbances on the measurement sensors. Norizuki et al. [25] made use of an MPC algorithm in a teleoperation system with a certain time delay, using a variable damping method on the master side for collision prevention. In order to reduce the consequences of the time delay on the general performance and stability of a surgical robot during a beating-heart surgery, Bowthorpe et al. [2] exploited an MPC algorithm to get rid of the delays of the position data acquisition of the beating heart from delayed ultrasound images.

The purpose of this study is to design an online robust model predictive control structure for the VR haptic simulator, which accounts for the assignment of the dynamic models to each of the aforementioned three major components of the systems. To that end, we proposed a phenomenological tissue fracture model to represent the dynamics of the virtual environment, that is, endoscopic sinus and skull base surgery [30-33]. The haptic interface, Novint Falcon (Novint Technologies, Inc.), is a 3-DOF parallel robot manipulator used in the experimental set-up in this research. We also equipped this robot with a calibrated force sensor at its end-effector to measure the exerted force from the operator's arm to the haptic interface. Moreover, due to the highly nonlinear dynamics of such a parallel manipulator, we divided the operational workspace of the robot's end effector into several sub-spaces, and piecewise linear models are designated to each of these operational sub-spaces. To characterize the human operator's arm dynamic model, a lumped five-parameter single-degree-of-freedom mass-stiffness-damping impedance model is utilized [5, 6]. Additionally, in the present work, we proposed an online robust model predictive control algorithm and studied the closed-loop stability of the entire VR haptic training simulator. To indicate the contributions of this paper, an improved approach for the quasi-min-max output feedback MPC for implementation on a linear parametric variable (LPV) as well as hybrid systems is addressed based on [23, 26]. With regard to the quasi-min-max algorithm, Park *et. al.* introduced an online robust output feedback MPC algorithm for linear parametric variable systems. Their suggested approach guaranteed the robust stability of the output feedback constrained systems. Some assumptions made in the development of our new approach are likewise to those of the work of Park *et. al.* [26]. However, in our proposed modified theory as opposed to the [26], by taking the characterized dynamic model into account, the one step ahead prediction horizon: $x_{k+1}$ as the next future states, can be acquired using the predictive dynamics. Moreover, we employed a cost function that uses $\Delta u$ instead of $u$, resulting in more robustness in spite of the presence of the switching phenomenon or disturbances. The proposed control algorithm implementation is novel in terms of control application in surgical simulators and is promising in the general applications of teleoperation or VR haptic systems. Furthermore, we tried to provide a relatively precise resemblance between the impedance of the virtual environment and the impedance that the human operator feels, rather than the impedance simply generated by the robot manipulator, which can further improve the system transparency.

To sum up, we can express the main contributions of this paper as follows: 1) Proposing a novel dynamic model for the sinonasal tissue to be applied to the virtual environment, which is derived and characterized by experimental tests; 2) Developing a VR-based haptic training system formulated as a linear parametric variable problem under the input constraints; 3) Using a novel output-feedback control algorithm to reduce or even eliminate the unwanted disturbances generated by the switching between the dynamics of the control effort signals; 4) Reduction of the consequences of the probable uncertainties in the prediction dynamics, that leads to the noticeable improvements in the system's robustness as well as the convergence of the system outputs to the desired impedance quantities.

The simulations and experiments are conducted to analyze the effectiveness of the proposed model and control algorithm, using a one-degree-of-freedom VR-based haptic training simulator. Through an empirical evaluation, the proposed modified MPC algorithm for the linear parametric variable problems can guarantee the robust stability of the input-constrained virtual reality-based haptic system in practical applications.

This paper is organized as follows: The components of the virtual-reality-based haptic system, the modified output feedback control for the linear parametric variable problem of the haptic system, and the control design and implementation for a single-axis example of the virtual haptic system are introduced and discussed in Sections 2 and 3, respectively. The analytical and experimental results are reported and compared in Section 4, and finally, the discussions and conclusions are drawn in Section 5.

## 2. The Components of the Virtual-Reality Based Haptic System

An ideal VR-based medical/surgical haptic training simulator should employ a physical model of the virtual environment capable of providing repeatable results comparable to the real environment in a closed-loop control framework and must be able to render both temporary and permanent changes in the simulated graphics of the respective virtual environments [27]. Moreover, developing an online robust model predictive control for such a simulator requires an prior knowledge of the dynamic model of the human operator, the haptic interface, and the virtual environment (Figure 1).

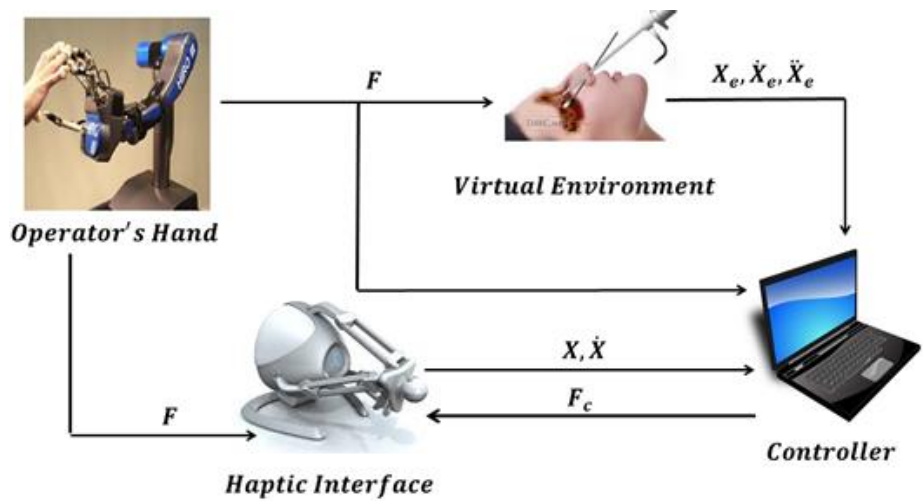

**Fig. 1** The schematic view of a haptic virtual system.

The co-existence of the soft and hard tissues in the sinus area results in the consequent complexity and nonlinearity of the mechanical behavior of this region, therefore, the modeling of which can be quite challenging. This challenge in the model characterization can be even more arduous if the tissue fracture phenomenon is considered in the simulation. Hence, an appropriate dynamic model formulation is indispensable, in advance of the human-robot interaction simulation. Therefore, a nonlinear indentation-rate-dependent model is utilized to represent the tissue behavior before and after the fracture. The proposed model is dependent on the tool's velocity and displacement during indentation into the tissue [32]. In 2019, the authors of this research [32] proposed a phenomenological fracture model for sinus tissue (Figure 2) based on the experimental data. The mechanical properties of the sinonasal regions were examined using sheep heads. The tool-tissue insertion scenarios were characterized by considering the effects of the indentation rate under various indentation and relaxation experiments. The tool insertion procedure into the coronal orbital floor (COF) tissue, under different indentation rates, was modeled as a sequence of deformation, fracture, and cutting stages.

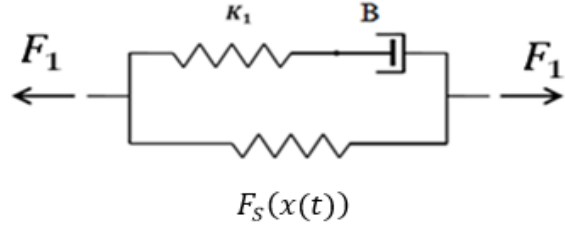

**Fig. 2** Nonlinear rate-dependent model of sinus tissue [1].

The aforementioned study introduced a better sense of tissue fracture during the surgery operation while presenting a complete tool-tissue interaction model. Their proposed nonlinear tissue behavior model prior and posterior to the fracture is considered as:

$$F(x, \dot{x}, t) = \begin{cases} F_1(x, \dot{x}) & x \leq X_f \\ F_2(x, \dot{x}) & x > X_f \end{cases}, \quad X_f(t) = H(\dot{x}), \qquad (1)$$

where $F_1$ is the nonlinear indentation-rate-dependent loading force on the tissue; and $F_2$ is the cutting force. $X_f$ is the boundary condition where the tissue fracture occurs. The functions presented in (1) are elaborated on as follows:

$$F_1 = F_s(x(t)) + K(x)v\tau_s\left(1 - exp\left(-\frac{x}{v\tau_s}\right)\right),$$

$$F_2 = F_f(v) + a(v) * \left(x_{post} - x_f(v)\right), \qquad (2)$$

$$x_f(v) = 0.0001v^2 - 0.0575v + 19.21$$

where $F_1$ is the classical standard nonlinear indentation-rate-dependent model derived from the modified Kelvin-Voigt (KV) model, in which $F_s(x(t))$ is the static function of the displacement, and the second term in the right side of the first line of (2) stands for the dynamic function of both displacement and tool-tissue interaction velocity ($v$) which is presented by a set of nonlinear springs and dampers, referred to a relaxation time constant ($\tau_s$). $x_f(v)$ is a criterion to determine the fracture occurrence in the COF. Additionally, $F_2$ represents the mechanical model of tissue cutting forces. $F_f(v)$ denotes that the fracture force depends on the rate of the tool indentation, and $x_{post}$ is the tool displacement after the fracture has occurred (for more details please read [28]). Table 1 and Table 2 describe expressions included in (2) and the constant values of the relaxation time constant, respectively.

**Table 1** Algebraic expressions involved in the nonlinear tissue behavior

| Parameter | Algebraic expression |
|---|---|
| $F_s(x(t))$ | $0.008x^3 + 2.087x^2 + 8.766x$ |
| $\tau_s$ | $B/K$ |
| $F_f(v)$ | $0.001v^2 - 1.176v + 697.1$ |
| $a(v)$ | $10^{-7}v^4 + -7 \times 10^{-5}v^3 + 0.0101v^2 + 0.0485v + -79.313$ |

**Table 2** Constant parameters of the relaxation time expression

| Parameter | $K$ | $B$ |
|---|---|---|
| Value | 63.62 | 0.021 |

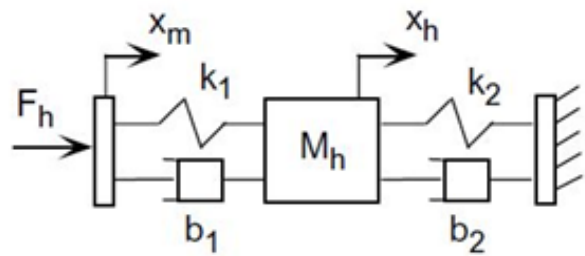

**Fig. 3** Representation of the five-parameter model for the hand dynamic.

The human operator's arm dynamic can be represented as a one-axis five-parameter impedance model [5] (Figure 3). The results presented in [5] demonstrate that the five-parameter dynamic model of the operator's arm was able to properly model the average behavior of several subjects with satisfactory precision. Table 1 reports the average values for the five-parameter dynamic model of the human arm derived from several experiments. These values will be used as an initial guess for the estimation of the human arm's actual five-parameter dynamic model in the experiments on a one-DOF haptic virtual system. As the parameters of the human arm dynamic model differ from one operator to another, the prior step to any experiment is to specifically estimate the human operator arm dynamic model parameters during interaction with the haptic interface and update the controller accordingly, aiming to improve the system's transparency and robustness. To that end, a recursive normalized least square algorithm with a forgetting factor is employed for online parameter identification of the operator arm impedance model [5].

Table 3 Characteristics of the five-parameter model for the hand dynamics.

| Parameters | M (Kg) | K1 (N/m) | K2 (N/m) | B1 (N.s/m) | B2 (N.s/m) |
|---|---|---|---|---|---|
| Values | 0.89 | 24.06 | 42.93 | 5.60 | 14.04 |

Given the mean quantities of the parameters presented in Table 3, and considering them as the initial values for the model updating procedure, the estimated parameters of this model can properly converge to the actual parameters. In the next step, by adjusting these parameters in the control loop, the results of the VR-based haptic simulator of the endoscopic sinus surgery will be examined. Due to the presence of nonlinearity in most dynamic systems, a piecewise linear approach can be utilized to represent the nonlinearities in a system and to reduce the analytical and computational efforts of the control design. Characterizing the dynamic model of the haptic interface is a necessary step for developing a virtual haptic training simulator. In this research, the Novint Falcon - a parallel impedance-type robot - [17] is utilized as the haptic interface device (Figure 4).

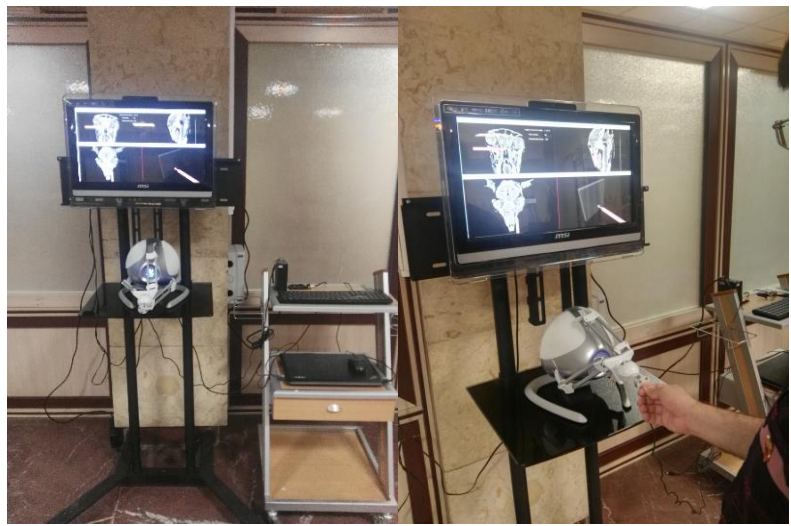

Fig. 4 The training system setup of the endoscopic sinus surgery.

Given that the Falcon robot can only set the position of its end-effector for the operator, a force sensor (UMMA5Kg.f load cell, Dacell Co.) is attached to the robot end-effector to measure the applied forces by the operator. A PC platform executes the control algorithm. An amplifier (DN-AM100, Dacell Co.) and an ARDUINO board with the baud rate value of 9600 bits per second (bps) are used for reading the sensor data via a serial port.

The dynamics of a haptic device in a single-axis coordinate ($X$) is modeled as [15]:

$$\begin{cases} \ddot{x} + 16.2\dot{x} = 2.02F, & x < -10mm \\ \ddot{x} + 22\dot{x} = 1.916F, & -10mm \leq x \leq 10mm \\ \ddot{x} + 37.1\dot{x} = 1.289F, & x > 10mm \end{cases} \quad (3)$$

## 3. Design of the Modified Model Predictive Control

### 3.1. System Description

Consider a multi-inputs multi-outputs discrete-time LPV system, presented by the dynamic model in the following equation:

$$x(k+1) = A(p(k))x(k) + B(p(k))u(k),$$
$$y(k) = C(p(k))x(k). \quad (4)$$

in which, $x(k) \in R^{n_x}$ is the unmeasurable states of the system, $y(k) \in R^{n_y}$ is the measurable output, $u(k) \in R^{n_u}$ is the control action, and $p(k)$ is the time-varying switching parameter, measurable at each step $k$, which relies on the measurable output of the system dynamics. It is assumed that $[A(p(k)), B(p(k)), C(p(k))] \in \Omega \equiv C_0\{[A_i, B_i, C_i] \ i = 1, 2, 3\}$ varies within a corresponding polytope $\Omega$. The vertices of the polytope consist of $L_g$ local system matrices which represent the number of piece-wise linear regions and are known at each step $k$. Moreover, $C_0$ denotes the convex hull of the aforementioned polytope. Since the states of the system are unmeasurable, an off-line robust observer is employed based on [37]. The following state observer is proposed to estimate the uncertainties in the states of the system in (4) as:

$$\hat{x}(k+1) = A(p(k))\hat{x}(k) + B(p(k))u(k)$$
$$+ L_o\left(y(k) - C(p(k))\hat{x}(k)\right), \quad (5)$$

in which, $\hat{x}(k)$ is the estimated state of the system and $L_0$ is the gain of the designed observer which is necessary to be determined in terms of linear matrix inequality (LMI). Considering [37], there exists a $P_0 > 0$ and $Y_0 = P_0 L_0$ which satisfies the following constraint equation:

$$\begin{bmatrix} \rho^2 P_0 - L_0 & P_0 A_j - Y_0 C \\ P_0 A_j - Y_0 C & P_0 \end{bmatrix} > 0, \tag{6}$$

which is affine in $\left[\left(A(p(k+i))\big|B(p(k+i))\right)\right]$ and $P_0$ is a positive definite matrix. The estimation error of the system is defined as $e(k) = x(k) - \hat{x}(k)$. Considering (4) and (5), the error of the dynamic system can be defined as:

$$e(k+1) = \left(A(p(k)) - L_0 C(p(k))\right) e(k), \tag{7}$$

By definition of a quadratic function for the estimation error $e(k)$, the obtained observer gain $L_0$ and the stability of dynamic error introduced as $E(e(k)) = e(k)^T P_0 e(k)$, $P_e > 0$, can be guaranteed for any $\left[\left(A(p(k))\big|B(p(k))\right)\right] \in \Omega$ and finally, for $k \to \infty$, one can conclude that $\hat{x}(k) \to x(k)$.

### 3.2. Online Modified Robust Output Feedback MPC

We are seeking a control algorithm design which stabilizes an LPV system while satisfying the input constraints and restating a tracking problem as a regulation one. Given the dynamics of the system in (4), it is supposed that the current system matrices $\left[\left(B(p(k))\right)\right]$ are known at step $k$. At the same time, the future ones $\left[\left(B(p(k+i))\right)\right]$ for $i \geq 1$ is uncertain with respect to the defined polytope $(\Omega)$. For a linear parameter variable system in (4), one can define a constraint as:

$$|u_j(k)| < u_{j,max}, \quad j = 1, 2, \ldots, n_u, \quad k \geq 0, \tag{8}$$

where, $|.|$ is the absolute value, and $u_{j,max}$ denotes the given upper bound on the $j^{th}$ input $u_j(k)$. The optimization problem can be represented as:

$$max J J_1^\infty < \delta(k), \tag{9}$$

where $\delta(k)$ is a non-positive value for the upper bound of the infinite horizon cost function $J_0^\infty(k)$, which can minimize the optimization problem, resulting in the calculation of the $u(k)$ as the control input, and $\Psi(k)$ for the future control law $u(k+i|k) = \Psi(k) x_p(k+i|k)$, $i \geq 1$ for $x_p$ as the predicted state of $x(k+i)$. Therefore, the optimization problems lead to:

$$\min_{u(k|k), Y(k), \Phi(k)} \delta(k), \tag{10}$$

Implementing the strategy in [39] results in the linear matrix inequalities as follows:

$$\begin{bmatrix} \eta & * & * & * \\ x_{p_k} & \Phi(k) & * & * \\ ((I+A)^T Q(I+A))^{1/2}\hat{x}(k) & 0 & \delta(k)I_{nx*nx} & * \\ (B^T QB + R)^{1/2}\Delta u(k) & 0 & 0 & \delta(k)I_{nu*nu} \end{bmatrix} \tag{11}$$
$$> 0,$$

$$\begin{bmatrix} \Phi(k) & * & * & * \\ S_j(k) & \Phi(k) & * & * \\ Q^{1/2}\Phi(k) & 0 & \delta(k)I_{nx*nx} & * \\ R^{1/2}Y(k) & 0 & 0 & \delta(k)I_{nu*nu} \end{bmatrix} > 0, \quad j = 1,2,\ldots,L_g \tag{12}$$

$$\delta(k)I - \varepsilon\Phi(k) > 0, \tag{13}$$

$$\begin{bmatrix} u_{j,max} & u_j(k) \\ u_j(k) & u_{j,max} \end{bmatrix} > 0, \tag{14}$$

$$\begin{bmatrix} Su(k) & * \\ Y(k)^T & \Phi(k) \end{bmatrix} > 0, \quad \text{and } Su_{jj} < u_{j,max}^2, \quad j = 1,2,\ldots,n_u \tag{15}$$

$Su$ is a symmetric matrix and (11) and (12) are axisymmetric matrices and $\eta$ and $S_j(k)$ are introduced as:

$$\begin{aligned} \eta &= 1 + [L_o(y - C_k\hat{x}_k) + u(k-1)]^T Q[L_o(y - C_k\hat{x}_k) \\ &\quad + u(k-1)], \\ S_j(k) &:= A_j\Phi(k) + B_jY(k), \end{aligned} \tag{16}$$

In our model predictive control theory, for the future states prediction of the system, presented by (4), the following predictive model is considered:

$$\begin{aligned} x_p(k+1+i|k) &= A(p(k+i))x_p(k+i|k) \\ &\quad + B(p(k+i))u(k+i|k) \\ x_p(k|k) &= \hat{x}(k), \quad i \geq 1 \end{aligned} \tag{17}$$

in which, $x_p(k+i|k)$ denotes the predicted states of $x(k+i)$ of the system of (1) and $u(k+i|k)$ is the future control input for step $k+i$ estimated at step $k$, respectively. For this purpose, assume that $x_p(k|k) := \hat{x}(k)$ and the first element of the computed control sequence is $u(k|k)$ and in the future steps, $u(k+i|k)$ can be followed by the following equation:

$$u(k+i|k) = \Psi(k)x_p(k+i|k), \quad i \geq 1 \tag{18}$$

where, $\Psi(k)$ is the feedback gain introduced at step k. Improving the optimization strategy for introduced by [39], this method minimizes the infinite horizon objective function $J_0^\infty(k)$ at each step $k$ as:

$$\min_{u(k+i|k)} \max_{\left[\left(A(p(k+i))\big|B(p(k+i))\right)\right]\in\Omega, i\geq 0} J_0^\infty(k), \quad (19)$$

$$J_0^\infty(k) = J_0^1(k) + J_1^\infty(k),$$

Equation (19) denotes that the control input $u(k)$ will satisfy the input constraints of the system revealed by (8). In (19), the terms are represented as:

$$\begin{cases} \Delta u(k) = u(k), & k = 1 \\ \Delta u(k) = u(k) - u(k-1), & k > 1 \end{cases}$$

$$J_0^\infty(k) = \sum_{i=0}^{\infty} \{x_p(k+i|k)^T Q x_p(k+i|k)$$

$$+ \Delta u(k+i|k)^T R \Delta u(k+i|k)\},$$

$$J_0^1(k) \coloneqq x_p(k|k)^T Q x_p(k|k) + \Delta u(k|k)^T R \Delta u(k|k),$$

$$J_1^\infty(k) = \sum_{i=1}^{\infty} \{x_p(k+i|k)^T Q x_p(k+i|k)$$

$$+ \Delta u(k+i|k)^T R \Delta u(k+i|k)\},$$

$$J_1^\infty(k) \coloneqq \sum_{i=1}^{\infty} \{x_p(k+i|k)^T Q x_p(k+i|k) \quad (20)$$

$$+ \Delta u(k+i|k)^T R \Delta u(k+i|k)\}$$

$$= x_p(k+1|k)^T Q x_p(k+1|k) + JJ_1^\infty$$

$$JJ_1^\infty = \sum_{i=2}^{\infty} \{x_p(k+i|k)^T Q x_p(k+i|k)\}$$

$$+ \sum_{i=1}^{\infty} \{\Delta u(k+i|k)^T R \Delta u(k+i|k)\},$$

where $R > 0$ and $Q > 0$ are the weighting matrices. Introducing the upper bound on $JJ_1^\infty$, which is quite different from what is proposed by [41, 33], the quasi-min–max optimization problem for MPC can be derived. Therefore, the following proposed function is assumed:

$$V\left(x_p(k+i|k)\right) = x_p(k+i|k)^T \Gamma(k) x_p(k+i|k), \quad i > 0, \Gamma(k) > 0 \quad (21)$$

It is supposed that $V\left(x_p(k+i|k)\right)$ satisfies the following stability condition at each step k for $\left[\left(A(p(k+i))\big|B(p(k+i))\right)\right] \in \Omega, i \geq 1$. Then, one can briefly write the following equation:

$$V_{k+i} = x_{p_{k+i}} \Gamma_k x_{p_{k+i}} \tag{22}$$

Considering (22) for any future step results in:

$$V_{k+i+1} - V_{k+i} < -J_{k+i}^{k+i+1},$$
$$J_{k+i}^{k+i+1} = x_{p_{k+i}} Q x_{p_{k+i}} + \Delta u_{k+i}^T R \Delta u_{k+i}, \tag{23}$$

For the robust stability of a system, it is necessary that $V(\hat{x}(\infty|k)) = 0$. Considering (23), it is necessary to define the cost function explicitly in all future steps of stability functions of $V_k$ and $V_{k+1}$. Then, a robust control rule can be obtained which is optimized for all future steps. Taking into account $V(k)$, we can define:

$$J_0^\infty(k) < V_k, \tag{24}$$

in which,

$$JJ_1^\infty + x_{p_{k+i}}^T Q x_{p_{k+i}} + J_0^1 < V_k, \tag{25}$$

or in summary

$$JJ_1^\infty < V_k - x_{p_{k+i}}^T Q x_{p_{k+i}} - J_0^1, \tag{26}$$

In (26), the left side of the inequality is indefinite, and the right side shows the maximum of $JJ_1^\infty$. If u(k|k), Y(k) and $\Phi$(k) are available for $\hat{x}$(k) and y(k), the feedback gain $\Psi$(k) can be determined from $\Psi(k) := Y(k)\Phi(k)^{-1}$. Equations (8)-(15) refer to the functional conditions of (9) and the stability condition of (23). If there exist $Y(k)$, $\Phi(k)$, and $\delta(k)$ such that (10) holds, and also if the following inequality relation is true:

$$\hat{x}_p(k|k)^T Q \hat{x}_p(k|k) + \Delta u(k|k)^T R \Delta u(k|k)$$
$$+ \hat{x}_p(k+i|k)^T \Gamma(k) \hat{x}_p(k+i|k) < \delta(k), \tag{27}$$

Then:

$$\hat{x}_p(k+i|k)^T \Gamma(k) \hat{x}_p(k+i|k) < \delta(k), \tag{28}$$

Assuming that $Q \ \& \ R > 0$, considering (23), then the following relation can be derived:

$$\hat{x}_p(k+i+1|k)^T \Gamma(k) \hat{x}_p(k+i+1|k)$$
$$- \hat{x}_p(k+i|k)^T \Gamma(k) \hat{x}_p(k+i|k)$$
$$< -[\hat{x}_p(k|k)^T Q \hat{x}_p(k|k)$$
$$+ \Delta u(k|k)^T R \Delta u(k|k)], \quad i \geq 1 \tag{29}$$

That is

$$\hat{x}_p(k+i+1|k)^T \Gamma(k) \hat{x}_p(k+i+1|k)$$
$$< \hat{x}_p(k+i|k)^T \Gamma(k) \hat{x}_p(k+i|k), \quad i \geq 1 \tag{30}$$

Therefore, $\varepsilon := \{\hat{x}_p \in R^{n_x} | \hat{x}_p^T \Gamma(k) \hat{x}_p < \delta\}$ is an invariant ellipsoid for the predicted future states $\hat{x}_p(k+i|k)$, $i \geq 1$, of the uncertain system. Based on (27)-(30), the input constraint can be cast as LMI constraints as presented in (13)-(15). To guarantee the robust stability of the system at step k, we first need to instate that the optimization problem in (10) can be a feasible solution to the problem at step k+1. For this purpose, we set that:

$$\Delta u(k+1|k+1) = \Delta u(k+1|k) = \Psi(k)\hat{x}_p(k+i|k),$$
$$\Gamma(k+1) = \Gamma(k), \tag{31}$$

Then, it is necessary to check (11), (13), and (14), since they include $s(k)$, $\hat{x}(k)$, $y(k)$, and $u(k|k)$, which are changed at each step k. First, by the satisfaction of (15) at step k, $\Delta u(k+1|k)$ is feasible. Then, considering $\Delta u(k+1|k+1) = \Delta u(k+1|k) = \Psi(k)\hat{x}_p(k+i|k)$, it is clear that (14) is satisfied at step k + 1. In the next step, we check (11). Based on (11) and (27)-(30), we substitute that $\Delta u(k+1|k+1) = \Delta u(k+1|k) = \Psi(k)\hat{x}_p(k+i|k)$ and $\Gamma(k+1) = \Gamma(k)$. Then, (11) can be rewritten as

$$\hat{x}_p(k+1|k)^T \zeta(k) \hat{x}_p(k+1|k)$$
$$- \hat{x}_p(k+1|k)^T \hat{\xi}(k+1)^T \Gamma(k) \hat{\xi}(k \tag{32}$$
$$+ 1) \hat{x}_p(k+1|k) < 0,$$

in which $\zeta(k) := \Gamma(k) - Q - F(k)^T R F(k)$ and $\hat{\xi}(k+1) = A(s(k+1)) + B(s(k+1))F(k)$. Since at step k, (12) is satisfied, the presented inequality in (32) holds. Therefore, the feasibility of the optimization problem in (10) is guaranteed.

Based on (12) and (13), there exists $\varepsilon > 0$ satisfying $\delta(k)I - \varepsilon\Phi(k) > 0$ for any $t > k$. Now, the stability constraint (27) guarantees that $\hat{x}_p(k|k) := \hat{x}(k)$, $u(k|k)$, and $\hat{x}_p(k+1|k)$ converge to zero, since $Q \& R > 0$, and $\Gamma(k) > \varepsilon I$. In addition, the estimated state $\hat{x}(k)$ will converge to the system state $x(k)\ for\ k \to \infty$. Therefore, the feasibility of solving the optimization problem (10) guarantees the asymptotic robust stability, (i.e., $x(k) \to \infty$ as $k \to \infty$). Then, the presented robust output feedback MPC scheme is introduced as $\min_{u(k|k),Y(k),\Phi(k),X(k)} \delta(k)$ subject to (11)–(15), (17) and (18) at each step k, for implementing the first control input $\Delta u(k|k)$ and $u(k|k)$, the optimal control sequence $u(k+i|k)$ will be determined by solving the aforementioned optimization, considering the measured output $y(k)$ and the time-varying parameter $s(k)$.

### 3.3. Control Implementation for a Single-Axis Virtual Haptic System

Guaranteeing the stability of the system as well as providing the required transparency of it are the two main necessities of an appropriate controller in such VR haptic simulators (Figure 5).

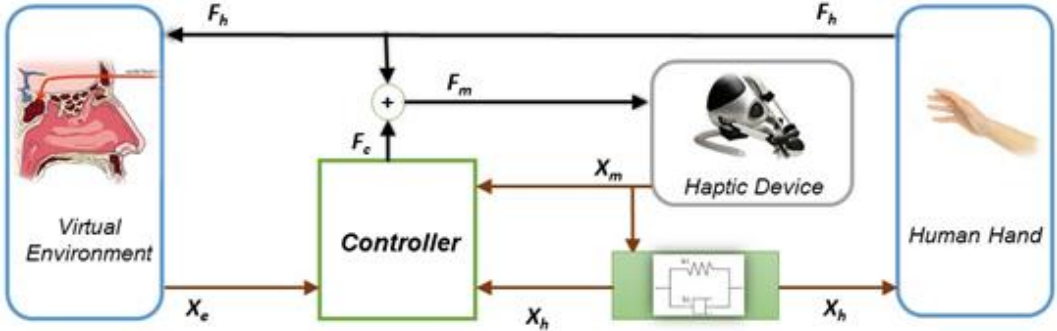

**Fig. 5** The schematic view of the components of a closed-loop haptic virtual system.

To provide realistic simulations, it is attempted to estimate the impedance of the operator's arm and to follow the desired impedance of the dynamic model of the virtual environment. This feature is another novelty of this paper. Hence, the convergence of the impedance of the operator's arm to that of the virtual environment is the desired goal of the control framework. To do so, we described the operator's hand displacement as a function of the control input. Given the operator arm dynamic model discussed in [5, 6], the utilized operator's arm force-displacement transfer function is represented as follows:

$$\frac{x_h}{F_c} = \frac{b_i}{S(S + a_i)} \frac{b_{h1}S + k_{h1}}{m_h S^2 + (b_{h1} + b_{h2})S + k_{h1} + h_{h2}}, \quad (33)$$
$$i = 1, 2, 3$$

in which, $x_h$ is the human arm's displacement, $F_c$ is the control effort, $m_h$ is the mass of the human arm, $b_{h1}$ and $b_{h2}$ are the damping coefficients and $k_{h1}$ and $k_{h2}$ are the stiffness coefficients of the arm. To eliminate the pole located at the origin, we restated the transfer function in the velocity format such that:

$$\frac{v_h}{F_c} = \frac{b_i}{(S + a_i)} \frac{b_{h1}S + k_{h1}}{m_h S^2 + (b_{h1} + b_{h2})S + k_{h1} + h_{h2}}, \quad (34)$$
$$i = 1, 2, 3$$

where, $v_h$ is the operator's hand velocity. Then, in a switching system, the displacement of the user interface ($x_m$), varies between the different piecewise linear regions. The state-space model is then discretized with the sample time of ($T_s = 0.005s$) to be used in the control structure. The zero-order-hold sampling technique is utilized for the system dynamics discretization. The discretized format of (4) which is both controllable and observable is restated as follows:

$$A_1 = 10^{-3} \times \begin{bmatrix} 879.7 & -64.3 & -42.61 \\ 150.3 & 994.8 & -3.482 \\ 6.141 & 79.86 & 999.9 \end{bmatrix},$$

$$B_1 = 10^{-3} \times \begin{bmatrix} 2.349 \\ 0.1919 \\ 0.005171 \end{bmatrix}, \quad C_1 = [0 \quad 0.3891 \quad 0.2637],$$

(35)

$$A_2 = 10^{-3} \times \begin{bmatrix} 853.1 & -71.38 & -57.73 \\ 148.1 & 994.2 & -4.74 \\ 6.141 & 79.86 & 999.9 \end{bmatrix},$$

$$B_2 = 10^{-3} \times \begin{bmatrix} 2.314 \\ 0.19 \\ 0.005133 \end{bmatrix}, \quad C_2 = [0 \quad 0.3691 \quad 0.2502],$$

(36)

$$A_3 = 10^{-3} \times \begin{bmatrix} 789.9 & -88.16 & -93.57 \\ 142.7 & 992.8 & -7.779 \\ 5.934 & 79.81 & 999.8 \end{bmatrix},$$

$$B_3 = 10^{-3} \times \begin{bmatrix} 2.230 \\ 0.1854 \\ 0.00504 \end{bmatrix}, \quad C_3 = [0 \quad 0.2483 \quad 0.1683],$$

(37)

## 4. Results

### 4.1. Robust Observer Performance Result

The initial states of the system are set to $x(0) = [0 \quad 0 \quad 0.001]^T$, the initial states of the observer are set to $\hat{x}(0) = [0 \quad 0 \quad 0.0015]^T$, the decay rate is $\rho = \sqrt{0.7}$, and $L_e = I_{3 \times 3}$ is considered as the weighting matrix. We used MATLAB to solve the LMI problem of (14) which is resulted in (38):

$$P_e = 10^8 \begin{bmatrix} 0.1004 & -0.3294 & -0.2415 \\ -0.3294 & 3.3957 & 2.3057 \\ -.02415 & 2.3057 & 1.6150 \end{bmatrix}, \quad Y_e$$

$$= 10^8 \begin{bmatrix} -0.0804 \\ 4.6058 \\ 3.1403 \end{bmatrix},$$

(38)

$$L_p^T = [-54.4612 \quad 253.5053 \quad -368.1313],$$

A step function with the amplitude of 5N is applied to the system as the input force, to examine the performance of the off-line state observer. Figure 6 shows the estimated system output and Figure 7 describes the off-line state observer error in the tracking of the output signals, respectively. The acquired results demonstrate that the designed observer not only estimates the states with accuracy, but it shows a great level of robustness while the switching phenomenon takes place in the dynamic of the system as well.

## 4.2. Experimental Result

To draw a comparison between the controller performance in both the simulation program and the real case, the previously recorded data sets of the experimental hand forces were implemented in a simulation program in MATLAB. In order to evaluate the performance of the designed controller with the sampling time step of $(T_s = 0.005s)$ the input constraints are set to $|u_c| \leq 10 - F_s$, $L = 1.5. I_{4\times4}$, $R = 1$, and $\varepsilon = 0.001$. At each step, for determining the reference signal, we read the force sensory data to determine the interaction force and to generate the proper control action to be applied by the robot for more efficient tracking purposes.

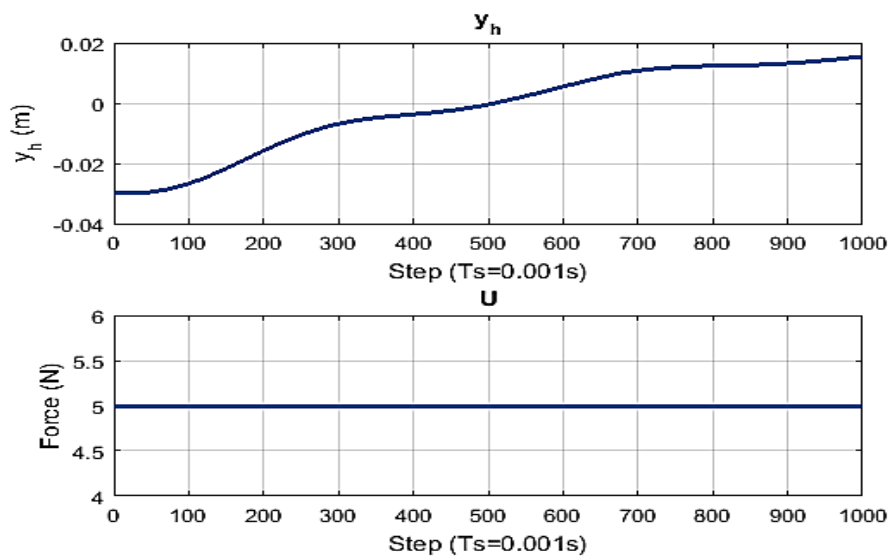

Fig. 6 The estimated system output calculated by the input force as a step function of amplitude 5N

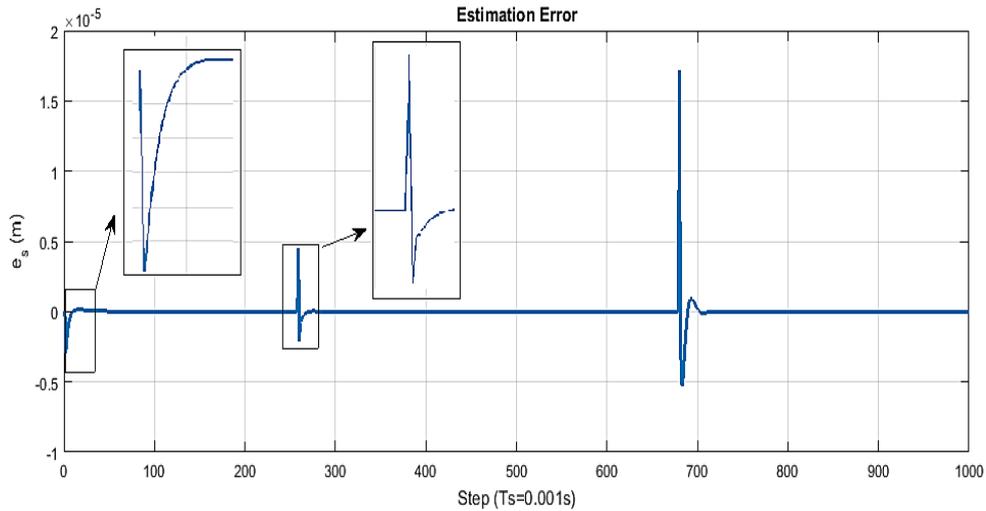

**Fig. 7** The off-line state observer error in the tracking of the output signal.

Recruiting the users for running the experimental setup, we recorded several interacting forces to generate the tracking reference from the tissue virtual environment. To properly adjust the control parameters, we run a simulated system in MATLAB considering both the aforementioned forces and the generated tracking signals. Hereafter, the performance of the haptic surgical system in simulating the fracture behavior of the tissue is compared both in the MATLAB simulation and in the experimental setup, with the same forces of the human operator. In the procedure of using a surgical simulator, different operators experienced the process of tissue fracture in a virtual environment. Each user experienced tissue fracture with different indentation rates, and the interaction forces are measured by the force sensor and recorded. These forces and the displacements of the robot's end-effector have been compared to the simulation results. Subsequently, the associated forces of three different human operators who have experienced the fracture of the sino-nasal tissue are depicted (Figure 8). Given the difference in their hand dynamics, hand-eye coordination, reaction time, and skills, they may perform the task at different time intervals. According to the general trend of the force diagram in Figure 8, we can divide its behavior into three major phenomenological phases. First, the tool pushes the coronal orbital floor surface up to a certain deformation such that a linear increment in the force quantity can be noticed until a slight drop occurs in the force magnitude (phase 1). Second, as the applied force reaches that certain deformation, the tool will penetrate abruptly into the coronal orbital floor by crossing the surface of the tissue (this is the reason for the slight drop of the force). After this point, the force quantity shall rise with a nonlinear but smooth pattern until it reaches the fracture point, beyond which a sharp abrupt drop is observed (phase 2). Third, as the force reaches the fracture point, the cutting phase of the tissue begins and will be continued by inserting the tool further into the damaged tissue (phase 3).

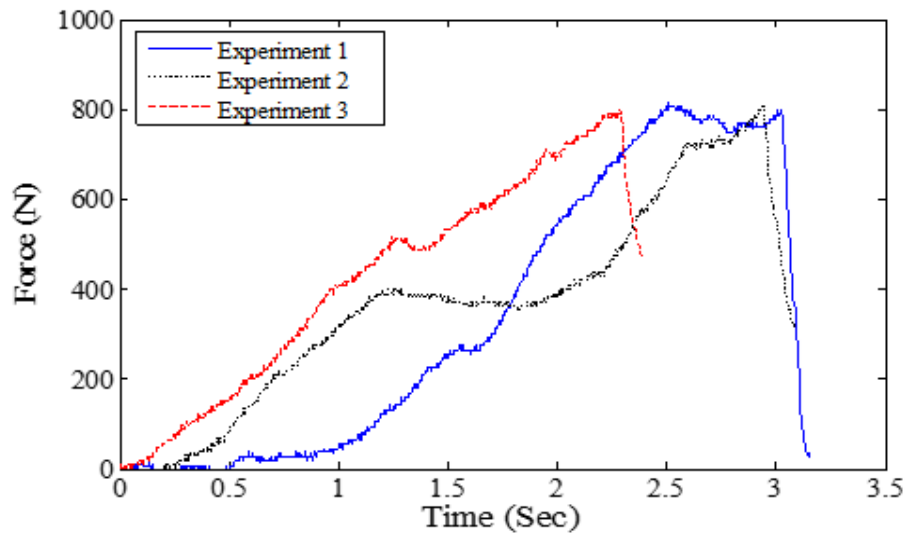

**Fig. 8** The three different human hand and robot end effector interaction forces.

The tracking performance and error of the robot's end-effector with respect to the displacement as well as the impedance are described and analyzed both in the simulations and experiments. Figure 9 illustrates the comparison between the actual displacement of the robot's end-effector, the simulated signal, and the desired reference signal. As Figure 9 shows, the VR-based haptic system, either in the simulation or in the experiment phase, will effectively meet the control purpose. In spite of promising results, there are small deviations involved in simulation results for achieving control purposes (Figure 9-a-c). Computation delays, local optimization calculations, and the switches between the dynamics of the model and the robot's actuators saturation after repeated usage can be regarded as the main sources of the possible deviation. Although some deviations occurred in the acquired results, the objectives of the control structure and the virtual reality-based system are well achieved.

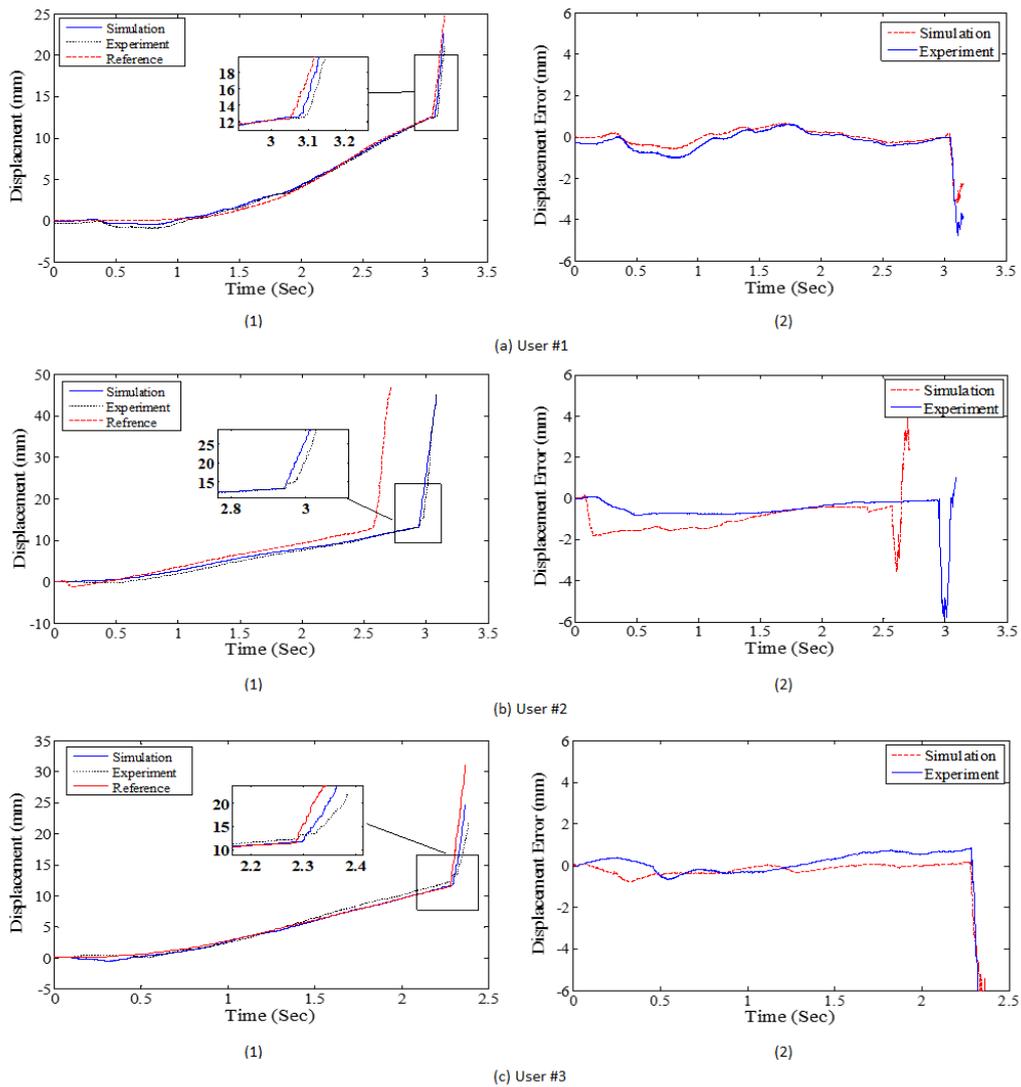

**Fig. 9** (1) The reported displacement of the reference signal created by the tissue dynamic model, the effective robot displacement in both simulation and experiment and (2) the tracking error of different input forces from different users.

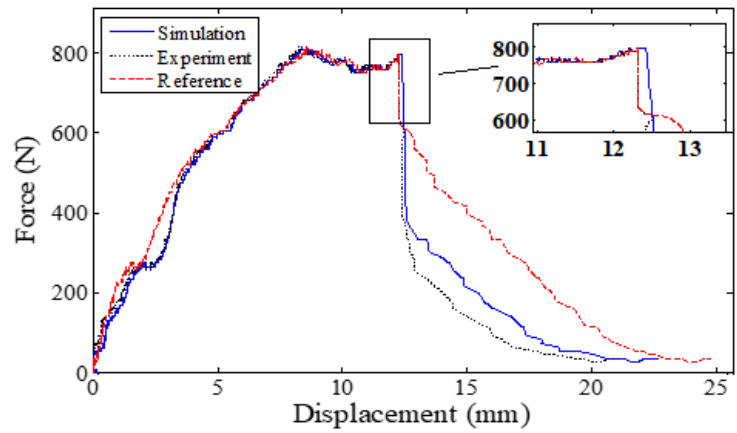

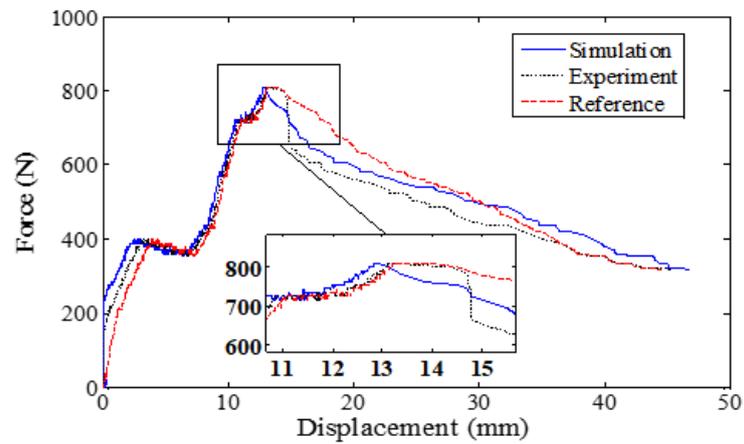

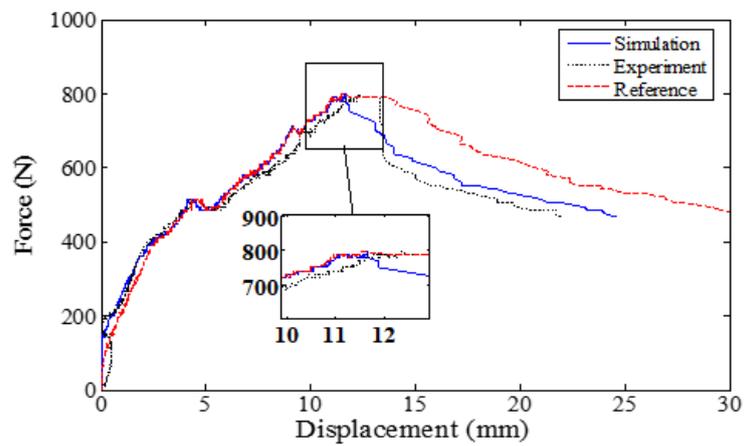

**Fig. 10** The comparison of the desired reference impedance of tissue, simulated impedance, and actual impedance.

The control objective is to properly track the displacement in the virtual environment. Figure 10 shows the desired reference impedance of the tissue in comparison to the simulated and actual signals. According to Figure 10, it can be seen that the haptic system is effectively following the impedance of the virtual environment in both the simulations and the experiments. Consequently, it leads to an effective surgery simulation in terms of transferring an actual haptic sense. Figure 10 shows that the acquired simulation and experiment results properly follow the desired dynamics both prior to the tissue fracture point and posterior to that point, while some slight deviations in the presented results can be observed. This phenomenon could be due to a sudden change in the reference path, in robot actuators functionality, and an unintended abrupt effort from the operator to manipulate the robot end effector position. The mentioned reasons may decrease the quality of the robot and the controller's performance after the fracture point, as expected in the simulation. However, despite the deviation increment, the robot demonstrates a similar behavior to that of the tissue's mechanical model. This means that the sense of tissue fracture, the maximum fracture force, as well as the ensuing sudden changes of the force associated with large displacements, can be well simulated or transmitted to the operator's hand. Table 2 reports the percentage of the root mean square error of the control tracking purpose for the data acquired from the experiments with respect to the ones from the simulations. Accordingly, the RMSD (root mean square deviation) of error for the obtained data shows a variation of 1.5% to 2.9% in the simulations and 2.1% to 3.9% in the experiments.

Table 4 RMSD (root mean square error deviation) of experimental data with respect to the ones from a simulation in tracking the objective of the control scheme.

| Indentation rate | Trial 1 | | Trial 2 | | Trial 3 | |
|---|---|---|---|---|---|---|
| | Simulation | Experiment | Simulation | Experiment | Simulation | Experiment |
| RMSD of error (%) | 1.5 | 2.1 | 2.5 | 3.2 | 2.9 | 3.9 |

5. Conclusion

In this paper, we developed a haptic virtual reality-based Endoscopic Sinus Surgery (ESS) training simulator and simulated a virtual environment for the sinonasal tissue in CHAI3D open software and proposed a novel dynamic model for it, which is derived and characterized by experimental force-displacement data collected from real sheep heads in the operation room. Then we implemented a quasi-min-max output feedback model predictive control on the haptic training system. The haptic system subjected to the input constraint is modeled as a linear parametric variable system. Using the proposed output-feedback control approach and employing a new type of cost function, the proposed MPC approach is successful in reducing the unwanted disturbances caused by switching in the piecewise linear dynamics of the control effort signals. This property makes the proposed approach more convenient for implementation on nonlinear systems in practice. Therefore, the effects of the potential uncertainties in the prediction of the system's dynamics are reduced, which results in the improvement of the system's robustness and the convergence of the output signals to the desired impedance values. The analytical and experimental results demonstrated that a satisfactory tradeoff between performance and stability can be accomplished. A thorough validation study of the proposed ESS training system is presented in [33].

## 6. Acknowledgment

We thank the Djavad Mowafaghian Research Center for Intelligent NeoruRehabilitation Technologies of the Sharif University of Technology and the Bio-Inspired System Design Laboratory of the Biomedical Engineering Department at Amirkabir University of Technology (Tehran Polytechnic) for supporting this research. In addition, we are grateful to Ehsan Abdollahi for helping us to prepare the experiment set-up system.

## 7. Funding

This research did not receive any specific grant from funding agencies in the public, commercial, or not-for-profit sectors.

## 8. Conflicts of Interest Statement

The authors confirm that this manuscript and its corresponding research work involve no conflict of interest.

## 9. Ethical Approval

Not required

**References**

[1] Amirkhani, G., Farahmand, F., Yazdian, S. M., Mirbagheri, A.: An extended algorithm for autonomous grasping of soft tissues during robotic surgery. Int. J. Med. Robot. Comput. Assist. Surg. **16**(5), 1-15 (2020). https://doi.org/10.1002/rcs.2122

[2] Bowthorpe, M., Tavakoli, M.: Generalized predictive control of a surgical robot for beating-heart surgery under delayed and slowly-sampled ultrasound image data. IEEE Robot. Autom. Lett. **1**(2), 892-899 (2016). https://doi.org/10.1109/LRA.2016.2530859

[3] Choi, K. S., He, X., Chiang, V. C. L., Deng, Z.: A virtual reality based simulator for learning nasogastric tube placement. Comput. Biol. Med. **57**, 103-115 (2015).

https://doi.org/10.1016/j.compbiomed.2014.12.006

[4] De Rossi, G., Muradore, R.: A bilateral teleoperation architecture using Smith predictor and adaptive network buffering. IFAC-PapersOnLine **50**(1), 11421-11426 (2017).

https://doi.org/10.1016/j.ifacol.2017.08.1806

[5] Esfandiari, M., Sadeghnejad, S., Farahmand, F., Vosoughi, G.: Adaptive characterisation of a human hand model during intercations with a telemanipulation system. IEEE 3rd RSI Int. Conf. Robot. Mechatron. (ICROM) (pp. 688-693) (2015).

https://doi.org/10.1109/ICRoM.2015.7367866

[6] Esfandiari, M. and Farahmand, F.: Emg-based neural network model of human arm dynamics in a haptic training simulator of sinus endoscopy. In *2021 IEEE Int. Conf. Robot. Autom. (ICRA)* (pp. 1790-1795) (2021). https://doi.org/10.1109/ICRA48506.2021.9561555

[7] Esfandiari, M., Sadeghnejad, S., Farahmand, F. and Vosoughi, G.: Robust nonlinear neural network-based control of a haptic interaction with an admittance type virtual environment. IEEE 5th RSI Int. Conf. Robot. Mechat. (ICROM) (pp. 322-327) (2017).

https://doi.org/10.1109/ICRoM.2017.8466196

[8] Faulwasser, T., Findeisen, R.: Nonlinear model predictive control for constrained output path following. IEEE Trans. Automat. Contr. **61**(4), 1026-1039 (2015).

https://doi.org/10.1109/TAC.2015.2466911

[9] Golnary, F., Moradi, H.: Dynamic modelling and design of various robust sliding mode controls for the wind turbine with estimation of wind speed. Appl. Math. Model. **65**, 566-585 (2019). https://doi.org/10.1016/j.apm.2018.08.030

[10] Hannaford, B., Ryu, J. H.: Time-domain passivity control of haptic interfaces. IEEE Trans. Robot. Autom. **18**(1), 1-10 (2002). https://doi.org/10.1109/70.988969

[11] Harischandra, P. A., Abeykoon, A. M.: Upper-Limb Tele-Rehabilitation System with Force Sensorless Dynamic Gravity Compensation. Int. J. Soc. Robot., **11**(4), 621-630 (2019). https://doi.org/10.1007/s12369-019-00522-1

[12] Hokayem, P. F., Spong, M. W.: Bilateral teleoperation: An historical survey. Aut., **42**(12), 2035-2057 (2006). https://doi.org/10.1016/j.automatica.2006.06.027

[13] Jain, S., Lee, S., Barber, S. R., Chang, E. H., Son, Y. J.: Virtual reality based hybrid simulation for functional endoscopic sinus surgery. IISE Trans. Healthc. Syst. Eng. **10**(2), 127-141 (2020). https://doi.org/10.1080/24725579.2019.1692263

[14] Ji, Y., Gong, Y: Adaptive Control for Dual-Master/Single-Slave Nonlinear Teleoperation Systems With Time-Varying Communication Delays. IEEE Trans. Instrum. Meas. **70**, 1-15 (2021). https://doi.org/10.1109/TIM.2021.3075527

[15] Khadivar, F., Sadeghnejad, S., Moradi, H., Vossoughi, G.: Dynamic characterization and control of a parallel haptic interaction with an admittance type virtual environment. Meccanica **55**(3), 435-452 (2020). https://doi.org/10.1007/s11012-020-01125-1

[16] Khadivar, F., Sadeghnejad, S., Moradi, H., Vossoughi, G., Farahmand, F: Dynamic characterization of a parallel haptic device for application as an actuator in a surgery simulator. IEEE 5th RSI Int. Conf. Robot. Mechat. (ICROM) (pp. 186-191) (2017). https://doi.org/10.1109/ICRoM.2017.8466168

[17] Kolbari, H., Sadeghnejad, S., Bahrami, M., Ali, K. E.: Adaptive control of a robot-assisted tele-surgery in interaction with hybrid tissues. Journal of Dynamic Systems, Meas. Control. **140**(12) (2018). https://doi.org/10.1115/1.4040818

[18] Kolbari, H., Sadeghnejad, S., Bahrami, M., Kamali, A.: Bilateral adaptive control of a teleoperation system based on the hunt-crossley dynamic model. IEEE 3rd RSI Int. Conf. Robot. Mechat. (ICROM) (pp. 651-656) (2015).

https://doi.org/10.1109/ICRoM.2015.7367860

[19] Kolbari, H., Sadeghnejad, S., Bahrami, M., Kamali, E. A.: Nonlinear adaptive control for teleoperation systems transitioning between soft and hard tissues. IEEE 3rd RSI Int. Conf. Robot. Mechat. (ICROM) (pp. 055-060) (2015).

https://doi.org/10.1109/ICRoM.2015.7367760

[20] Lee, S.M., Kwon, O.M., Park, J.H.: Output feedback model predictive tracking control using a slope bounded nonlinear model. J. Optim. Theory Appl. **160**, pp.239-254 (2014). https://doi.org/10.1007/s10957-012-0201-8

[21] Lee, S.M., Won, S.C. and Park, J.H.: New robust model predictive control for uncertain systems with input constraints using relaxation matrices. J. Optim. Theory Appl. **138**, pp.221-234 (2008). https://doi.org/10.1007/s10957-008-9375-5

[22] Li, H., Zhang, L., Kawashima, K.: Operator dynamics for stability condition in haptic and teleoperation system: A survey. Int. J. Med. Robot. Comput. Assist. Surg. **14**(2), e1881 (2018). https://doi.org/10.1002/rcs.1881

[23] Lu, Y., Arkun, Y.: Quasi-min-max MPC algorithms for LPV systems. Automatica 36(4), 527-540 (2000). https://doi.org/10.1016/S0005-1098(99)00176-4

[24] Moreira, P., Zemiti, N., Liu, C., Poignet, P.: Viscoelastic model based force control for soft tissue interaction and its application in physiological motion compensation. Comput. Meth. Programs Biomed. **116**(2), 52-67 (2014). https://doi.org/10.1016/j.cmpb.2014.01.017

[25] Norizuki, H., Uchimura, Y.: Contact prediction control for a teleoperation system with time delay. IEEJ J. Ind. Appl. **7**(1), 102-108 (2018). https://doi.org/10.1541/ieejjia.7.102

[26] Park, J. H., Kim, T. H., Sugie, T.: Output feedback model predictive control for LPV systems based on quasi-min–max algorithm. Automatica **47**(9), 2052-2058 (2011). https://doi.org/10.1016/j.automatica.2011.06.015

[27] Piromchai, P.: Virtual reality surgical training in ear, nose and throat surgery. Int. J. Clin. Med. *5*(10), 558-566 (2014). https://doi.org/10.4236/ijcm.2014.510077

[28] Polushin, I. G., Liu, P. X., Lung, C. H.: A force-reflection algorithm for improved transparency in bilateral teleoperation with communication delay. IEEE/ASME Trans. Mechatron. **12**(3), 361-374 (2007). https://doi.org/10.1109/TMECH.2007.897285

[29] Rosseau, G., Bailes, J., del Maestro, R., Cabral, A., Choudhury, N., Comas, O., DiRaddo, R.: The development of a virtual simulator for training neurosurgeons to perform and perfect

endoscopic endonasal transsphenoidal surgery. Neurosurg. **73**(suppl_1), S85-S93 (2013). https://doi.org/10.1227/NEU.0000000000000112

[30] Sadeghnejad, S., Elyasi, N., Farahmand, F., Vossughi, G., Sadr Hosseini, S. M.: Hyperelastic modeling of sino-nasal tissue for haptic neurosurgery simulation. Sci. Iran. **27**(3), 1266-1276 (2020).

[31] Sadeghnejad, S., Esfandiari, M., Farahmand, F., Vossoughi, G.: Phenomenological contact model characterization and haptic simulation of an endoscopic sinus and skull base surgery virtual system. IEEE 4th Int. Conf. Robot. Mechatron. (ICRoM), pp. 84-89 (2016). https://doi.org/10.1109/ICRoM.2016.7886822

[32] Sadeghnejad, S., Farahmand, F., Vossoughi, G., Moradi, H., Hosseini, S. M. S.: Phenomenological tissue fracture modeling for an endoscopic sinus and skull base surgery training system based on experimental data. Med. Eng. Phys. **68**, 85-93 (2019).

https://doi.org/10.1016/j.medengphy.2019.02.004

[33] Sadeghnejad, S., Khadivar, F., Abdollahi, E., Moradi, H., Farahmand, F., Sadr Hosseini, S. M., Vossoughi, G.: A validation study of a virtual-based haptic system for endoscopic sinus surgery training. Int. J. Med. Robot. Comput. Assist. Surg. **15**(6), e2039 (2019). https://doi.org/10.1002/rcs.2039

[34] Sapkaroski, D., Baird, M., McInerney, J., Dimmock, M. R.: The implementation of a haptic feedback virtual reality simulation clinic with dynamic patient interaction and communication for medical imaging students. J. Med. Radiat. Sci. **65**(3), 218-225 (2018). https://doi.org/10.1002/jmrs.288

[35] Schwenzer, M., Ay, M., Bergs, T., Abel, D.: Review on model predictive control: An engineering perspective. Int. J. Adv. Manuf. Technol. **117**(5), 1327-1349 (2021). https://doi.org/10.1007/s00170-021-07682-3

[36] Seo, C., Kim, J. P., Kim, J., Ahn, H. S., Ryu, J.: Robustly stable bilateral teleoperation under time-varying delays and data losses: an energy-bounding approach. J. Mech. Sci. Technol. **25**(8), 2089-2100 (2011). https://doi.org/10.1007/s12206-011-0523-8

[37] Sirouspour, S., Shahdi, A.: Model predictive control for transparent teleoperation under communication time delay. IEEE Trans. Robot. **22**(6), 1131-1145 (2006).

https://doi.org/10.1109/TRO.2006.882939

[38] Song, A., Wu, C., Ni, D., Li, H., Qin, H.: One-therapist to three-patient telerehabilitation robot system for the upper limb after stroke. Int. J. Soc. Robot. **8**(2), 319-329 (2016). https://doi.org/10.1007/s12369-016-0343-1

[39] Sun, D., Naghdy, F., Du, H.: Application of wave-variable control to bilateral teleoperation systems: A survey. Annu. Rev. Control. **38**(1), 12-31 (2014). https://doi.org/10.1016/j.arcontrol.2014.03.002

[40] Tavakoli, M., Carriere, J., Torabi, A.: Robotics, smart wearable technologies, and autonomous intelligent systems for healthcare during the COVID-19 pandemic: An analysis of the state of the art and future vision. Adv. Intell. Syst. **2**(7), 2000071 (2020). https://doi.org/10.1002/aisy.202000071

[41] Torabi, A., Zareinia, K., Sutherland, G. R., Tavakoli, M.: Dynamic reconfiguration of redundant haptic interfaces for rendering soft and hard contacts. IEEE Trans. Haptics, **13**(4), 668-678 (2020). https://doi.org/10.1109/TOH.2020.2988495

[42] Uddin, R., Ryu, J.: Predictive control approaches for bilateral teleoperation. Annu. Rev. Control. **42**, 82-99 (2016). https://doi.org/10.1016/j.arcontrol.2016.09.003

[43] Vrooijink, G. J., Denasi, A., Grandjean, J. G., Misra, S.: Model predictive control of a robotically actuated delivery sheath for beating heart compensation. The Int. J. Robot. Res. **36**(2), 193-209 (2017). https://doi.org/10.1177/0278364917691113